# Orthographic Structuring of Human Speech and Texts: Linguistic Application of Recurrence Quantification Analysis

Franco Orsucci, Kimberly Walter*    Alessandro Giuliani†

Charles L. Webber Jr.‡    Joseph P. Zbilut§

November 2, 2018

## Abstract

A methodology based upon recurrence quantification analysis is proposed for the study of orthographic structure of written texts. Five different orthographic data sets (20th century Italian poems, 20th century American poems, contemporary Swedish poems with their corresponding Italian translations, Italian speech samples, and American speech samples) were subjected to recurrence quantification analysis, a procedure which has been found to be diagnostically useful in the quantitative assessment of ordered series in fields such as physics, molecular dynamics, physiology, and general signal processing. Recurrence quantification was developed from recurrence plots as applied to the analysis of nonlinear, complex systems in the physical sciences, and is based on the computation of a distance matrix of the elements of an ordered series (in this case the letters consituting selected speech and poetic texts). From a strictly mathematical view, the results show the possibility of demonstrating invariance between different language exemplars despite the apparent low-level of coding (orthography). Comparison with the actual texts confirms the ability of the method to reveal recurrent structures, and their complexity. Using poems as a reference standard for judging speech complexity, the technique exhibits language independence, order dependence and freedom from pure statistical characteristics of studied sequences, as well as consistency with easily identifiable texts. Such studies may provide phenomenological markers of hidden structure as coded by the purely orthographic level.

*Institute for Complexity Studies,American University, Rome, Italy

†TCE Laboratory,Istituto Superiore di Sanita, V.le Regina Elena 299 Rome, Italy

‡Department of Physiology, Loyola University, 2160 S. First Ave., Maywood, IL 60153 USA

§Department of Molecular Biophysics and Physiology, Rush University, 1653 W. Congress, Chicago, IL 60612 USA

# 1 Introduction

Mathematical language analysis has enjoyed varying degrees of success [14,8,5,7]. The use of orthography as a basis for computational studies of human languages has experienced much less enthusiasm owing to the view that orthography actually represents a superficial amalgam of phonological changes and conventions–an opinion which has had little objective confirmation [1].

In order to evaluate this view, a reliable metric is needed to compare different texts (orthographic sequences) corresponding to speech samples. This implies a technique able to identify and measure the amount of "prosodic structuring" of a given text. Important requirements the technique must meet include: 1) relative invariance with respect to the original language of the texts; 2) consistency with already known and recognizable prosodic structures; 3) maximal dependence upon dynamical (order dependent) features of texts coupled with a relative independence upon statistical (order independent) features. Furthermore, the chosen technique must be independent of distributional assumptions. The preliminary experimental evidence suggests that recurrence quantification analysis (RQA) meets these requirements.

# 2 Materials and Methods

## 2.1 Linguistic Data Sets

### 2.1.1 Italian poems of the XXth century (set 1)

Nine Italian poems of the last century were chosen with the aim of maximizing the range of variation of the degree of prosodic structuring. The texts include poems with a minimum level of redundancy, as well as very structured poems using long repetitive motifs. [CAPR1, G. Caproni: Il passaggio di Enea; SAB1, U. Saba: A mia moglie; GOZ1, G. Gozzano: La signorina Felicita; COM1, G. Comisso: Pesca miracolosa; MONT1, E. Montale: Meriggio; CAPR2, G. Caproni: Interludio; UNG1, G. Ungaretti: Sono una creatura; PALA1, A. Palazzeschi: Lasciatemi divertire; PALA2, A. Palazzeschi: Lasciatemi divertire (cont.). All poems from Carlo Salinari and Carlo Ricci (editors) (1976). *Storia della Letterature Italiana* (Vol. 3), Bari: Editore Laterza, except for Caproni's poems, which are from Giorgio Caproni (1968). *Il "terzo libro" e altre cose.* Torino: Giulio Einaudi.].

### 2.1.2 American (English) poems of the XXth century (set 2)

This set is made up of 11 American poems chosen with the same criteria adopted for set 1. [GIN1, A. Ginsberg: A vow; EAR1, E. A. Robinson: Richard Cory; WCW1, W. C. Williams: This is Just to Say; RF1, R. Frost: Revelation; RM1, R. Mezey: My mother; RF3, R. Frost: Stopping by woods on a snowy evening;

FOH1, F. O'Hara: Why I'm not a painter; EP1, E. Pound: To Em-Nei's "The unmoving cloud"; WS1, W. Stevens: Disillusiionment of ten o'clock; DS1, Dr. Seuss (pseudo.)(1960): *Green eggs and ham,* New York: Random House; RF2, R. Frost: The death of the hired man. All poems from George Moore (editor) (1977). *The Penguin Book of American Verse.* London: Penguin Books, except for the Frost poems, which are from Robert Frost (1915). *A boy's will and North of Boston* (1991 edition). New York: Dover].

### 2.1.3 Swedish contemporary poems and their correspondent Italian translations (set 3)

This set is made up of 12 Swedish poems together with their Italian translations. It is important to note that the translation was a "poetical" one, i.e. the translator had the goal to reproduce the prosodic structure of the original poem, so as to provide a check for the language invariance of the method. [TT1, T. Transtromer: Stenarna; TT2, T. Transtromer: Fran mars–79; GP1, G. Palm: Megafonen i poesiparken; GP2, G. Palm: Havet; GS1, G. Sonnevi: Bild i hostgront; GS2, G. Sonnevi: Vad formar; GT1, G. Tunstrom: Slutlig onskan; GT2, G. Tunstrom: Tva vindar; CA1, C. Anderson: Ibland far; CA2, C. Anderson: Skuggor; JW1, J. Werup: 29 Juli; JW2, J. Werup: Nyarsnatt. All poems and translations from *Poesia*. February, 1997, Vol. 101: 36–57. Rome: Crocetti editore].

### 2.1.4 Italian speech samples (set 4)

The transcriptions of the speech samples of 7 people with different cultural backgrounds. (3000-4000 letter lengths.)

### 2.1.5 American (English) speech samples (set 5)

The transcriptions of the speech samples of 10 students of the American University of Rome. The recorded speech samples correspond to periods of time in which the individuals are freely and fluently speaking with no external constraints and after a period of adaptation to the presence of the recorder.

## 2.2 Data analysis methods

RQA was first introduced in physics by Eckmann, et al. [2] Later, Webber and Zbilut [10] enhanced the technique by defining nonlinear variables that were found to be diagnostically useful in the quantitative assessment of time series structure in fields ranging from molecular dynamics to physiology [3,6,4,9,11,12,13]. RQA is based on the computation of a distance matrix between the rows (epochs) of an embedded matrix of the correspondent series (an embedding is a form of lagging to develop surrogate "dimensions"; see [10], for a detailed explanation of the method). Starting from this distance matrix, 5 nonlinear descriptors of the dynamics are evaluated. As pertains to this work, it is sufficient to describe the 2 basic descriptors: percent recurrence (REC) and percent determinism (DET). REC is the percentage of recurrent pairs of points in the above described distance matrix: a pair is considered recurrent if the distance between the elements is lower than a predetermined cutoff. In our case, given the symbolic character of the studied series, the cutoff was set to 0, and only totally superimposable epochs (text segments) in the embedding matrix are considered recurrent. This choice makes the analysis completely independent from the chosen code. DET is the percentage of recurrent points that appear in sequence forming diagonal line structures in the distance matrix. DET corresponds to the existence of patches of recurrent behavior in the studied series. In our application of the method, the texts were taken to be continuous streams of letters with no interruptions, with the letters being coded by alphabetical order. The embedding dimension was set to 3 in order to situate the window of analysis at the word level. Different choices of the embedding dimension (4 and 5 dimensions) gave coinciding results (data not shown). The choice of a 3 dimensional embedding maximizes the sensitivity of the method by increasing the number of scored recurrences without disturbance from low-level statistical features of the language (asymmetrical distribution of couples of letters typical of each language). A recurrence is scored whenever 2 identical three letter patterns are recognized along the sequence. The departure of the analyzed texts from random sequences was evaluated by comparing the real sequences with their randomly shuffled counterparts. The shuffling procedure keeps invariant the statistical features of the studied sequence so as to evaluate the amount of pure dynamical (i.e. phasic) information present in the text at the orthographic level.

## 3 Results

### 3.1 Sets 1 and 2 (Italian poems, Figs. 1 and 2)

It is worth noting the disposition of the poems going from maximum complexity (left-lower corner of the graph) to very recurrent, deterministic structures (right-superior comer). The "complex" end is made up of poems with no discernible rhyme or rhythmic structure (CAPR1, SAB1), while the low complexity end is occupied by texts with highly repetitive motifs (PALA1, PALA2). When the poems were shuffled, the randomized series with RQA profiles were hardly discernible from the most complex texts and very far apart from the highly structured poems. All the poems, when shuffled, gave very similar results [Mean (REC) = 0.35, SD = 0.01; Mean (DET = 8.00; SD = 0.32)], and were significantly different from their shuffled counterparts with the only exceptions being SAB1 and CAPR1. The high correlation between REC and DET variables (Pearson's r = 0.90, p < 0.01) highlights a common scaling of the poems in the REC-DET plane pointing to a common "structural design" of the texts, which was further confirmed by the next analyses.

The 2 planes globally coincide indicating a common general structure for the poems of both languages. (Fig. 2 reports both Set 1 and Set 2 data.) It is worth noting some peculiarities: 1) RF1 is the only American poem not statistically discriminable from its random counterpart (shuffled version) and, in analogy with the Italian "very complex" texts like CAPR1, does not present any noticeable redundancy. 2) The EAR1 and UNG1 poems occupy a very similar position in the REC-DET plane. Reading the texts makes apparent how the two poems are characterized by the same general structural feature; i.e., the iteration of a common pattern organizing the text. 3) The extremely high redundancy of Doctor Suess' poem (DS1) is immediately recognizable by the method and probably represents a real maximum of structuring degree as well as RF1. CAPR1 and SAB1, given their contiguity with random sequences, probably constitute a real minimum.

## 3.2 Set 3 (Swedish poems and Italian translations, Figs. 3-4)

The analysis of this data set allowed us to investigate the issue of the recognizability of a given prosodic structure across two different languages. The demonstration of the possibility to recognize the same structure (a poem) in terms of RQA parameters when translated into another language, is both a stringent test of the language invariance of the technique, as well as of the consistency of the dynamical descriptors. The obvious linguistic differences between Swedish and Italian languages makes the test more cogent. The Swedish texts were transliterated using the Italian 21 letter standard alphabet (ch for k, and so on) maintaining the same phonetic character of the original. Figures 3 and 4 report the statistically significant correlations between the Swedish and Italian versions (r = 0.85 and r = 0.90 for REC and DET respectively, $p < 0.01$). It is worth noting that the technique was able to discover the prosodic linkage between the texts without being affected by the obvious morphological differences between Italian and Swedish languages.

## 3.3 Set 4 (Italian speech samples, Fig. 5)

It is first important to note how human speech samples (SP) fit very well in the REC-DET plane individuated by poems maintaining the strong relation of the two descriptors (r = 0. 89, $p < 0.001$). This suggests that the speech samples have a general structure similar to the poems. While the prosodic structure of the poems is characterized by very short sequences, the same general structure in speech samples involves more lengthy texts (3000-4000 letters). This fact makes the poems the ideal reference standard for speech samples because they give condensed (and thus easily understandable) specimens of the prosodic structure representative of long speech samples. In the case of the Italian speech samples, the most representative poem is CAPR2 which scales in the area of Fig. 5 occupied by the speech samples and in fact presents a spoken language-like organization. The speech samples are significantly different from their shuffled counterparts thus highlighting an orthographic structuring.

### 3.4 Set 5 (American speech samples, Fig. 6)

In this case too there is a correlation between the 2 descriptors ($r = 0.88$, $p < 0.001$), confirming the fit of the spoken language samples into the space spanned by the poems. The homogeneity of the speakers is reflected by the tight distribution of the samples, and the American and Italian samples have a similar scaling: [American Mean (REC) = 0.416, SD = 0.03, Mean (DET) = 20.00, SD = 1.36; Italian Mean (REC) = 0.676, SD= 0.06; Mean (DET) = 27.94, SD = 2.66 ]. Nevertheless the 2 groups of speakers were significantly different for both REC and DET mean values (t-test corrected for variance non-homogeneity: $p < 0.0001$) pointing to a greater complexity (lesser degree of structuring) of American speech. This result is probably linked to a higher cultural background of the American population with respect to the Italian one, but further investigation is needed to resolve this point. The American speakers displayed less variance than the Italian ones, demonstrating a statistical significance for REC (F-test for homogeneity of variances: $F = 4.69$, $p < 0.04$), pointing to the possibility of using the method to discriminate sets of speakers. Figure 7 exhibits the combined analysis of American poems (AMP), Italian poems (ITP), American speech (AMS), and Italian speech (ITS). It is worth noting the common scaling of the texts along a linear relationship between REC and DET, ( $r = 0.87$, $p < 0.001$). This common scaling points to the possibility of using the position on the REC-DET plane as a simple numerical index of the relative complexity of a text.

## 4 Conclusion

The RQA technique provides a quantitative description of text sequences at the orthographic level in terms of structuring, and may be useful for a variety of linguistics-related studies.

## 5 Software

Extensive documentation, and the programs used for the analysis developed by CLW and JPZ can be obtained at http://homepages.luc.edu/ ~ cwebber. Original texts are available from Dr. Giuliani upon request.

# 6  Acknowledgments

The authors thank C. Manetti, Chemistry Dept., Univ. of Rome "La Sapienza", for software support; Prof. G. Smith, American Univ. of Rome, for material help; and Dr. C. Iacobini, Linguistics Dept., Univ. of Ferrara, for encouragement.